\documentclass[10pt,twocolumn,letterpaper]{article}

\usepackage{iccv}
\usepackage{times}
\usepackage{epsfig}
\usepackage{graphicx}
\usepackage{amsmath}
\usepackage{amssymb}
\usepackage{multirow}
\usepackage[switch]{lineno}
\usepackage[dvipsnames]{xcolor}
\usepackage{enumitem}

\usepackage[breaklinks=true,bookmarks=false]{hyperref}

\iccvfinalcopy 

\def\iccvPaperID{****} 

\ificcvfinal\fi

\begin{document}

\title{Gradient-based Sampling for Class Imbalanced Semi-supervised Object Detection}

\author{Jiaming Li$^{1,2}$\thanks{Equally-contributed authors.} \thanks{Work done during an internship at Baidu.}\quad
Xiangru Lin$^{2*}$\quad
Wei Zhang$^{2}$\quad
Xiao Tan$^{2}$\quad 
Yingying Li$^{2}$\quad 
\\
Junyu Han$^{2}$\quad
Errui Ding$^{2}$\quad
Jingdong Wang$^{2}$\quad
Guanbin Li$^{1,3}$\thanks{Corresponding author.}\quad\\
$^{1}$School of Computer Science and Engineering, Sun Yat-sen University, Guangzhou, China\\
$^{2}$Department of Computer Vision Technology (VIS), Baidu Inc., China\\
$^{3}$Research Institute, Sun Yat-sen University, Shenzhen, China \\
{\tt\small lijm48@mail2.sysu.edu.cn, liguanbin@mail.sysu.edu.cn } \\
{\tt\small \{linxiangru,zhangwei99,tanxiao01,liyingying05,hanjunyu,dingerrui,wangjingdong\}@baidu.com}
}

\maketitle
\ificcvfinal\thispagestyle{empty}\fi

\begin{abstract}
    Current semi-supervised object detection (SSOD) algorithms typically assume class balanced datasets (PASCAL VOC etc.) or slightly class imbalanced datasets (MS-COCO, etc). This assumption can be easily violated since real world datasets can be extremely class imbalanced in nature, thus making the performance of semi-supervised object detectors far from satisfactory. 
    Besides, the research for this problem in SSOD is severely under-explored. 
    To bridge this research gap, we comprehensively study the class imbalance problem for SSOD under more challenging scenarios, thus forming the first experimental setting for class imbalanced SSOD (CI-SSOD).
    Moreover, we propose a simple yet effective gradient-based sampling framework that tackles the class imbalance problem from the perspective of two types of confirmation biases. 
    To tackle confirmation bias towards majority classes, 
    the gradient-based reweighting and gradient-based thresholding modules leverage the gradients from each class to fully balance the influence of the majority and minority classes. 
    To tackle the confirmation bias from incorrect pseudo labels of minority classes, the class-rebalancing sampling module resamples unlabeled data following the guidance of the gradient-based reweighting module.
    Experiments on three proposed sub-tasks, namely MS-COCO, MS-COCO $\rightarrow$ Object365 and LVIS, suggest that our method outperforms current class imbalanced object detectors by clear margins, serving as a baseline for future research in CI-SSOD. Code will be available at
    \url{https://github.com/nightkeepers/CI-SSOD}.

\end{abstract}

\section{Introduction}
Different types of objects in nature appear at different frequencies, and there is bound to be class imbalance in the dataset corresponding to object detection. 
Class imbalance refers to the scenario when a number of classes are over-represented (also known as majority classes), having more samples than others (also known as minority classes) in the dataset. Semi-supervised object detection (SSOD) utilizes both labeled and unlabeled data to improve the performance of an object detector. Although current SSOD methods have achieved promising performance on standard benchmark datasets (MS-COCO, PASCAL VOC, etc), these datasets are assumed to be class balanced or slightly class imbalanced. This underlying assumption can be easily violated when applying current SSOD methods to real-world scenarios where datasets are extremely class imbalanced in nature. For example, the sample distribution of corner cases in autonomous driving is scarce. As a result, the performance of these SSOD methods often suffers, especially for those minority classes. Besides, the research efforts dedicated to tackling the class imbalance problem in SSOD are severely under-explored.

In contrast, there are extensive literature on learning a class-balanced detector in supervised settings such as Long-tailed object detection (LTOD) and Few-shot object detection (FSOD). 
LTOD balances the influence of majority and minority classes by data re-sampling~\cite{feng2021exploring,wu2020forest} and class-balanced losses~\cite{tan2021equalization,li2022equalized}. 
Current FSOD methods transfer knowledge learned from the majority classes to the minority classes by fine-tuning~\cite{qiao2021defrcn,kaul2022label} or meta-learning-based strategies~\cite{li2021beyond,han2021query}.
However, both LTOD and FSOD methods are devised in a supervised manner and thus cannot fully exploit the potential information in unlabeled data, making them prone to overfit the limited labeled images of minority classes.
Besides, an increasing number of research efforts are dedicated to tackling the class imbalance problem in semi-supervised learning for the image classification task~\cite{wei2021crest,he2021rethinking,oh2022daso}. But simply applying these methods to SSOD generates inferior performance as manifested in our experiments in later sections. This is because the resampling and pseudo labeling strategies are tailored for image-wise annotations but not for box-wise annotations. Since there always exist multiple instances in an image, the resampling of pseudo labels on image-aspect enlarges the negative impact of incorrect pseudo labels and ``supervision collapse''\cite{doersch2020crosstransformers}. 
Current SSOD methods alleviate the class imbalance problem by using focal loss~\cite{liu2021unbiased,liu2022unbiased} or by applying class-wise thresholds~\cite{chen2022dense,li2022rethinking}.  Nevertheless, these methods fail to generate high-quality pseudo labels for minority classes in the extremely class imbalanced semi-supervised datasets due to the so-called confirmation bias  \cite{arazo2020pseudo}. The confirmation biases are mainly from two parts: (1) The model biased toward majority classes tends to predict pseudo labels that are also biased toward majority classes. Using these pseudo labels to train the detector reinforces detectors to produce more pseudo labels for majority classes and thus overfits to the biased pseudo labels. (2) The pseudo labels of minority classes are prone to be incorrect. Since the proportion of ground truth annotations for these classes is small, the incorrect pseudo labels easily dominate the training for these classes, thus making the detector fit with the incorrect information of these pseudo labels.

\begin{figure}[ht]
    \centering
    \vspace{-1mm}
    \includegraphics[clip,trim={0 0 0 0}, width=0.8\linewidth]{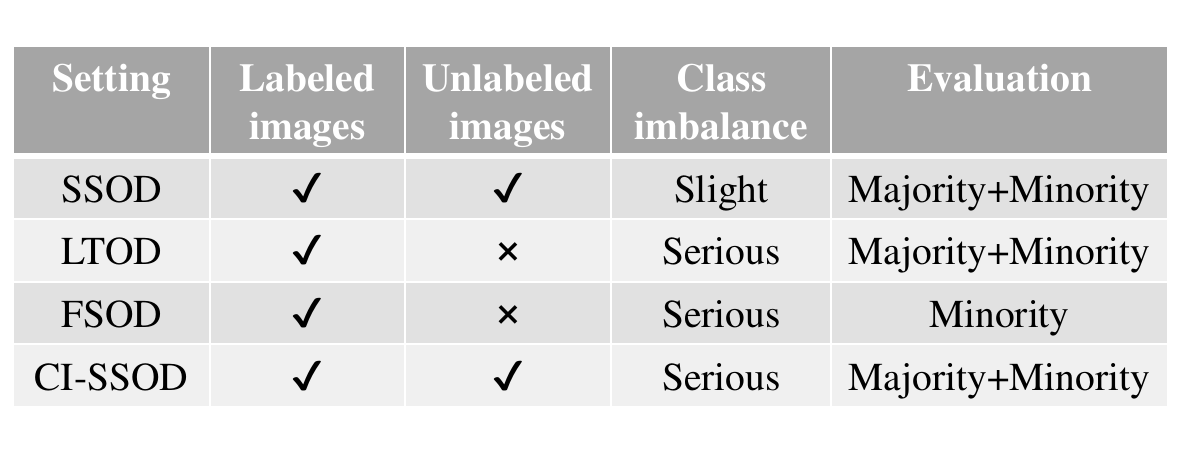}
    \vspace{-2mm}
    \caption{Differences of our proposed CI-SSOD with respect to other related object detection tasks. ``CI-SSOD'' represents class imbalanced SSOD. ``Majority'' and ``Minority'' represent the majority and minority classes respectively.}
    \vspace{-2mm}
    \label{fig:difference_task}
    \end{figure}
According to the analysis above, conducting a comprehensive study of the class imbalanced problem in SSOD requires considering two scenarios: (1) There are abundant unlabeled instances for the minority classes in the unlabeled images. A detector learned under such scenarios mainly suffers from the first type of confirmation bias commonly existed in current benchmarks. (2) The instances of minority classes in unlabeled images are naturally scarce. In this case, the second type of confirmation bias is the main obstacle to learn a balanced detector. To this end, we propose a new setting to comprehensively study the class imbalanced problem in SSOD, namely class imbalanced SSOD (CI-SSOD). Compared to existing benchmarks, CI-SSOD aims to build a balanced detector that could perform well for both majority and minority classes with the help of unlabeled data under more challenging class imbalance scenarios. We summarize the difference in Fig.~\ref{fig:difference_task}.

Designing a detector for CI-SSOD is still under-explored. Directly combining methods from LTOD (Eqlv2, etc.) or FSOD (DeFRCN, etc.) with SSOD (Soft teacher, etc.) generates inferior performance as proven in our experiments in later sections. The primary reason is that the combined methods fail to consider the two types of confirmation bias mentioned above, thus generating pseudo labels with low precision and recall and leading to ineffective usage of the unlabeled data. 
Monitoring dynamic model status during training is essential for designing a good detector under CI-SSOD. Based on this observation, we propose a simple yet effective gradient-based sampling framework for CI-SSOD from the perspective of two types of confirmation biases mentioned above.
To tackle the first type of confirmation bias, we propose a Gradient-based Reweighting (GbR) module that introduces a new perspective to fully balance class-wise positive and negative gradients by estimating class weights from a gradient matrix. This matrix serves as a metric to measure the model training status and formulates the optimization target for class weights, alleviating the confirmation bias toward majority classes.
Besides, we also present a Gradient-based Thresholding (GbT) module that modifies the class-wise thresholds for the unlabeled data following the guidance of the solved weights in the GbR module. 
To tackle the second type of confirmation bias, we present a Class-rebalancing Sampling (CrS) module. The CrS module considers not only class frequencies but also the confidence of pseudo labels and the dynamic class-wise thresholds from the GbT module, which drives the model to learn from pseudo labels with high confidence, alleviating the noise from incorrect pseudo labels.

To conclude, this paper has the following contributions:
\begin{itemize}
\item We extend the class imbalance study to SSOD, forming a new comprehensive setting, namely CI-SSOD.
\item We introduce a simple yet effective gradient-based sampling framework for CI-SSOD from the perspective of two types of confirmation biases. 
\item Extensive experiments on the MS-COCO, MS-COCO$\rightarrow$Object365, and LVIS sub-tasks show that our method outperforms all existing class imbalanced based methods by clear margins, demonstrating the superiority of our method.
\end{itemize}

\section{Related works}
\textbf{Semi-supervised object detection}. Current SSOD methods typically use consistency regularization or generate pseudo labels on unlabeled images~\cite{guo2022scale,chen2022dense,sohn2020simple,xu2021end,zhang2023sem,wang2023biased}. 
To overcome the class imbalance problem, Unbiased Teacher\cite{liu2021unbiased} and Unbiased Teacher-V2\cite{liu2022unbiased} adopt Focal loss\cite{lin2017focal} as the classification loss. \cite{li2022rethinking}, ~\cite{chen2022dense} and \cite{kaul2022label} adopt adaptive thresholds to sample pseudo labels based on the predicted scores of each class. \cite{guo2022scale} proposes a reweighting strategy to tackle the class imbalance between foreground and background regions. Although these mechanisms are conducive to alleviating the class imbalance problem, they are incapable of eliminating significant class imbalance in our proposed experiment setting, as demonstrated in our experiments. Our proposed gradient-based sampling framework is a tailored design for class imbalanced SSOD and can be easily adapted to different SSOD methods.

\textbf{Long-tailed object detection}. Current LTOD methods~\cite{li2022adaptive,feng2021exploring,zhang2021mosaicos} mainly focus on mitigating the class imbalance problem in object detection under the fully supervised setting. Resampling\cite{wu2020forest} is one of the main-stream methods to control the data distribution in a class imbalanced dataset. Repeat Factors Sampling(RFS)~\cite{gupta2019lvis} proposed to oversample the images of minority classes at the image level. 
Another type of solution is to adopt a balanced loss function\cite{li2020overcoming,ren2020balanced,wang2022c2am}.
EqlV2~\cite{tan2021equalization} and EFL~\cite{li2022equalized} reweight the loss function according to the gradients of each class.
They all suffer from overfitting minority classes and underfitting majority classes. Besides, these methods do not consider the confirmation biases \cite{arazo2020pseudo} commonly existed in semi-supervised learning and class imbalance in the unlabeled sets. Differently,  we estimate the model bias from the gradients imbalance perspective and tackle the different types of confirmation biases with a unified gradient-based sampling framework. This starkly contrasts current gradient-based methods in LTOD, which typically use gradient statistics as simple guidance and generate pseudo labels gradually biased toward majority classes.

\textbf{Few-shot object detection}. It aims to detect novel objects with only a few minority class annotations and abundant majority class annotations. Current approaches of FSOD can be roughly split into transfer-learning-based algorithms~\cite{qiao2021defrcn,kaul2022label} and meta-learning-based algorithms~\cite{li2021beyond,han2021query}. Different from our proposed experiment setting, FSOD does not provide unlabeled images. Besides, FSOD methods typically focus on the performance of minority classes, while CI-SSOD expects to enhance the performance of all classes.

\textbf{Class Imbalanced semi-supervised image classification}. There has been active research in solving the class imbalance problem in semi-supervised image classification. The core of these researches is based on a simple philosophy, which is sampling better minority-friendly pseudo label annotations from the unlabeled data~\cite{oh2022daso,lee2021abc,wei2021crest,he2021rethinking}. 
However, directly applying these methods to CI-SSOD generates inferior performance according to our experiments. The reason is that object detection contains many local background bounding boxes, which could introduce a large number of false negative noisy pseudo labels. This belongs to the second type of confirmation biases we discussed previously. We comprehensively analyze this confirmation bias and propose a class-rebalancing sampling strategy following the guidance of the estimated class-wise weights from the GbR module.

\section{Approach}
\begin{figure*}[ht]
    \centering
    \includegraphics[clip,trim={0 0 0 0}, width=0.9\linewidth]{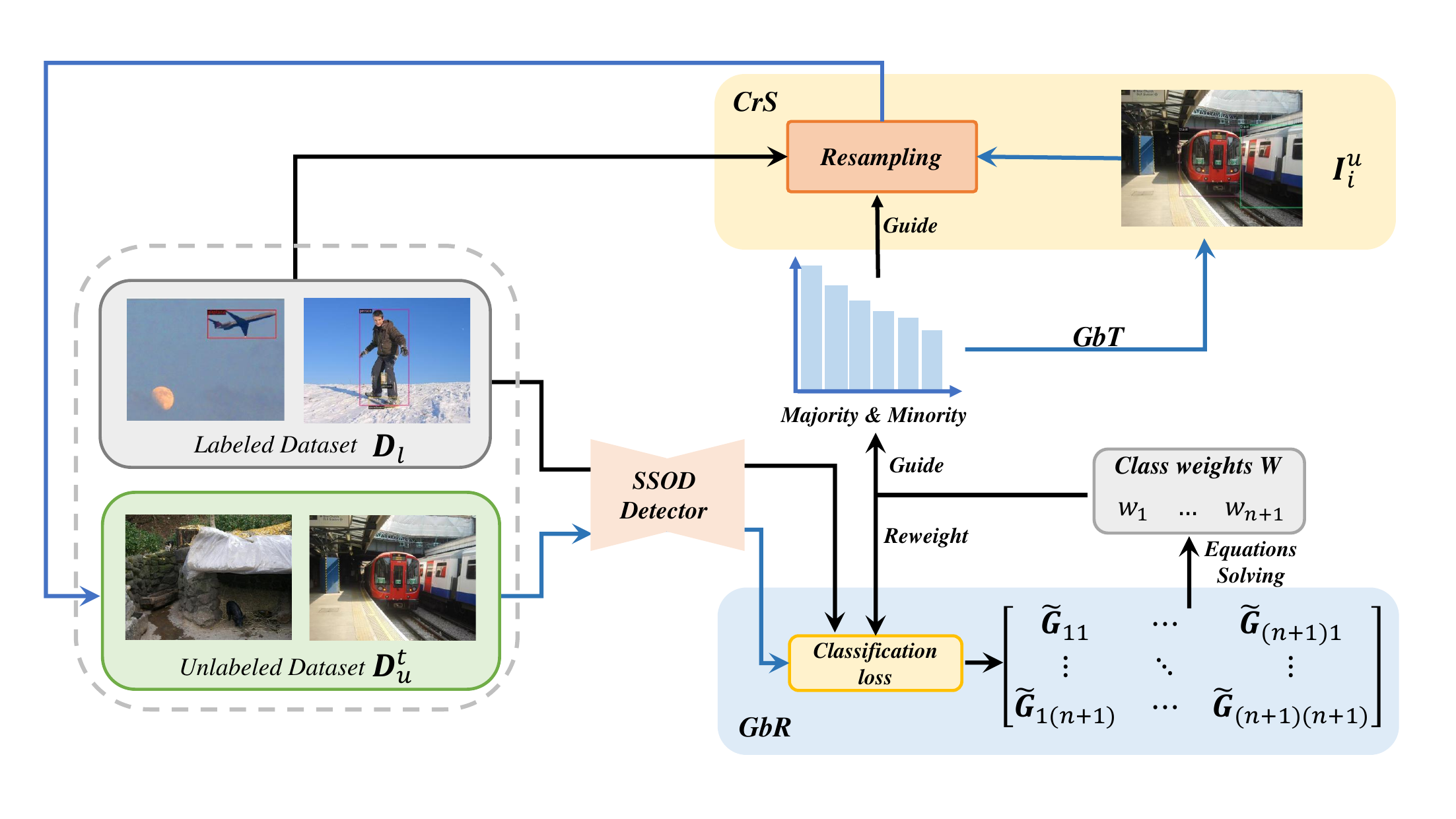}
    \vspace{-8mm}
    \caption{Overview of our proposed gradient-based sampling framework. Our framework consists of three main modules, namely the Gradient-based Reweighting (GbR) module, the Gradient-based Thresholding (GbT) module and the Class-rebalancing Sampling (CrS) module. The GbR module dynamically optimizes a set of graident-based linear equations to obtain the class-wise balancing weights $[w_1, ..., w_{n+1}]$. The GbT module guided by the class-wise balancing weights then performs adaptive thresholding to obtain class balanced pseudo labels from $\boldsymbol{D}_u$. The CrS module performs image-level data resampling on  $D_u$ with respective to the confidence and thresholds from GbT. Our method is agnostic to all SSOD methods.}
    \vspace{-3mm}
    \label{fig:pipeline}
    \end{figure*}

Our gradient-based sampling framework consists of three modules. The Gradient-based reweighting (GbR) fully balances class-wise positive and negative gradients by estimating class weights from a gradient matrix to tackle the first type of confirmation bias. The details are elaborated in Sec.~\ref{sec:GbR}. The Gradient-based Thresholding (GbT) modifies the thresholds based on weights from GbR to prevent the model from overfitting, which is elaborated in Sec.~\ref{sec:GbT}. To alleviate the influence of incorrect pseudo labels, class-rebalancing Sampling (CrS) resamples the unlabeled images based on the confidence of pseudo labels and thresholds in GbT. The details of CrS are further depicted in Sec.~\ref{sec:CrS}. Generally, the GbR and GbT modules help to eliminate the first type of confirmation bias and the CrS module alleviates the second type of confirmation bias. Our method mainly focuses on the classification branch of object detection, which suffers from a more serious imbalance than that of the localization branch.
Our method can be easily combined with current SSOD algorithms to tackle the class imbalance problem. An overview of our framework is presented in Fig.~\ref{fig:pipeline}.

\subsection{Problem Definition}
\label{sec:def}
To formalize the experiment setting, we define two types of data, namely the class imbalanced labeled data set as $\boldsymbol{D}_l = \{(\boldsymbol{I}^l_i, \boldsymbol{y}_{i}^{l})\}_{i=1}^{N_{l}}$  and the unlabeled set as $\boldsymbol{D}_u = \{(\boldsymbol{I}^u_i)\}_{i=1}^{N_{u}}$. 
Here $\boldsymbol{y}_{i}^{l} = \{(c_{j}^{l}, \boldsymbol{l}_{j}^{l})\}_{j=1}^{N_{i}}$ is a list of $N_{i}$ bounding box annotations composed of category labels $c_{j}^{l}$ and box location $\boldsymbol{l}_{j}^{l}$. $N_{l}$ is the number of labeled images, and $N_{u}$ is the number of unlabeled images. 
 Without loss of generality, we assume the classes as $\boldsymbol{C} = \{C_i \}_{i=1}^n$, where $n$ is the number of total classes. 
 
 We introduce two scenarios for the class-imbalanced SSOD experiment setting. \textbf{In the first scenario}, there are abundant unlabeled instances for the minority classes in the unlabeled images. To mimic the different levels of labeled data in the real world, the labeled set $\boldsymbol{D}_l$ can be split into a fully labeled data set for the majority classes $\boldsymbol{D}_m = \{(\boldsymbol{I}^l_i, \boldsymbol{y}_{i}^{l})\}_{i=1}^{N_{m}}$, a small amount of fully labeled data set for minority classes as $\boldsymbol{D}_s = \{(\boldsymbol{I}^l_i, \boldsymbol{y}_{i}^{l})\}_{i=1}^{N_{s}}$. Classes $\boldsymbol{C}$ can be split into the majority classes $\boldsymbol{C}^{f}= \{C_i\}_{i=1}^{n_f}$ and the minority classes $\boldsymbol{C}^{r} = \{C_i\}_{i=n_f+1}^{n_r+n_f}$.  Here  $n_f$ and $n_r$ are the numbers of majority classes and minority classes, respectively. Note that $n=n_f + n_r$. Obviously, the $\boldsymbol{D}_l$ and $\boldsymbol{D}_u$ do not share the same distribution.  \textbf{In the second scenario}, the instances of minority classes in unlabeled images are naturally scarce. In this scenario, the $\boldsymbol{D}_l$ and $\boldsymbol{D}_u$ share the same distribution. Since the $\boldsymbol{D}_u$ also suffers from extreme class imbalance, it is more challenging to design efficient pseudo labeling strategies for this scenario. 
 The goal of the task is to learn an object detector with $\boldsymbol{D}_l$ and $\boldsymbol{D}_u$ that can perform well on both majority and minority classes.

There are mainly two types of confirmation bias when applying existing methods to CI-SSOD: (1) the model biased toward majority classes tends to predict pseudo labels biased toward majority classes.  This type of confirmation bias refers to objects from minority classes being misclassified as majority classes or backgrounds.
(2) the incorrect pseudo labels for minority classes dominate the learning of the model since the ground truth labels in labeled data are scarce for these classes. This type of confirmation bias mainly arises when applying LTOD methods such as dynamic thresholding or resampling strategies in the second scenario.
In these cases, the objects from majority classes or backgrounds are annotated as pseudo labels of minority classes. As ground truth labels and the implicit instances of minority classes in unlabeled images are scarce, the learning of detectors is heavily interfered with by these incorrect pseudo labels. We further analyze the two types of confirmation bias in the Supplementary Document.

\subsection{Gradient-based Reweighting}
\label{sec:GbR}

We first analyze the reason for the performance degradation of directly combining the LTOD or FSOD methods with SSOD. 
For each class, we denote the gradients increasing the logits of this class as positive gradients and the gradients decreasing its logits as negative gradients. Concretely, for a given class, the gradients of all samples that possess the specific class label (or pseudo-label) are referred to as positive gradients, while the gradients of samples from other classes are referred to as negative gradients.
When positive gradients for a class overwhelm its negative gradients, the mean score of this class tends to become larger and the prediction of the detector could be biased toward this class. 
The higher diversity of majority classes leads to larger positive gradients than negative gradients for majority classes but vice versa for minority classes. Since the combined methods do not fully balance the gradients, the imbalance in gradients leads to confirmation bias toward majority classes. 
To alleviate this type of confirmation bias, we propose to force positive gradients and negative gradients for each class to be equal so that the logits of each class can level off during the training process.

Suppose the classification loss value on labeled images is $L^l$ and the classification loss value on unlabeled images is $L^u$.
We propose to reweight the classification loss $L^l = \sum_{j=1}^{N^l_b}{f^l(\boldsymbol{x}_j)}$ and $L^u = \sum_{j=1}^{N^u_b}{f^u(\boldsymbol{x}_j)}$ with class-wise weights based on its positive and negative gradients. The class logits of the $j$-th proposal in a batch produced by the classifier are denoted as $\boldsymbol{x}_j\in \mathbb{R}^{n+1}$, and the corresponding ground-truth label or pseudo ground-truth label is denoted as $c_j$. The $n+1$-th dimension of $\boldsymbol{x}_j$ is for the background class. $\boldsymbol{x}_j^i$ denotes the $i$-th logit,  which is corresponding to class $C_i$.
The $f^l(\cdot): \mathbb{R}^{n+1} \mapsto \mathbb{R}^{1}$ and $f^u(\cdot): \mathbb{R}^{n+1} \mapsto \mathbb{R}^{1}$ are the classification loss functions for labeled and unlabeled images. In this batch, $N^l_b$ and $N^u_b$ are the numbers of proposals in labeled images and unlabeled images, respectively.  

The positive gradient $\boldsymbol{G}_{ii}$ for class $C_i$ in a batch can be estimated by calculating the derivative of samples belonging to $C_i$(positive samples to $C_i$) to its logit $x_j^i$. Thus, for all positive samples, we have
\begin{equation}
    \boldsymbol{G}_{ii} = \sum_{\forall j, c_j = C_i}  \frac{\partial f^l(\boldsymbol{x}_j)}{\partial x_j^i} + \sum_{ \forall j, c_j = C_i}  \frac{\partial f^u(\boldsymbol{x}_j)}{\partial x_j^i}.
\end{equation}
By estimating the partial derivative to $x_j^i$ from the samples belonging to another class $C_k$(negative samples to $C_i$), the negative gradients $\boldsymbol{G}_{ik}$ for $C_i$ from $C_k$ in a batch are formulated as follows,
\begin{equation}
\boldsymbol{G}_{ik} =  \sum_{\forall j, c_j = C_k}  \frac{\partial f^l(x_j)}{\partial x_j^i} + \sum_{\forall j, c_j  = C_k}  \frac{\partial f^u(x_j)}{\partial x_j^i}.
\end{equation}
Note that  $\boldsymbol{G}_{ki} = 0$ for $k=1,2,...,n+1$ when there are no proposals whose category labels are $C_i$ in this batch.
For each batch, we obtain a gradients matrix $\boldsymbol{G}$ where the value on the $i$-th row and the $j$-th column is $\boldsymbol{G}_{ij}$. The $i$-th row represents the gradients to $C_i$ from proposals of each class and the $j$-th column represents the gradients from proposals of $C_j$ to each class.
To obtain a stable estimation of the gradient statistics, the final gradients are estimated by the moving average of gradients in a single batch, $     \boldsymbol{\tilde G} = \eta_g  \boldsymbol{\tilde G} + (1-\eta_g) \boldsymbol{G}.$
The positive gradients and negative gradients for each class are temporarily imbalanced, leading to a bias for the majority classes. To balance the positive and negative gradients for each class, we propose to reweight the classification loss with a set of learnable class-wise loss weights. For class $C_i$, the loss weight is defined as $w_i$. For each foreground class $C_i$, we expect the positive gradients and negative gradients to be equal. Therefore, we have the following equation, 
\begin{equation}
    w_i \times \boldsymbol{\tilde G}_{ii} = - \sum_{k \neq i} w_k \times   \boldsymbol{\tilde G}_{ik}.
\end{equation}

 As presented above, we generate $n+1$ equations regarding $w_i(i=1, ..., n+1)$. In order to keep the scale of weights unchanged, an additional equation is added to the aforementioned equations,
\begin{equation}
    \sum_{i=1}^{n+1}{w_i} = n+1.
    \label{eql:sum}
\end{equation}
We replace the $n+1$-th row representing the background class gradients with the above Equation~(\ref{eql:sum}). 

Finally, we optimize the set of linear equations defined above to obtain the final class-wise weights $w_i(i=1, ..., n+1)$, which are then leveraged to reweight the classification loss function to achieve class balance training. 
We smooth the weights $w_i^l = w_i^{\beta}$ on the labeled set $\boldsymbol{D}_l^t$ by a factor $\beta \in (0,1)$ to prevent the model from overfitting.

The traditional direct solutions to optimize the set of linear equations work well for tasks with relatively small numbers of classes. However, when samples of some minority classes are rarely sampled in the training process, the estimated weights are sometimes negative value or even do not exist. To tackle it, we utilize an optimizer to simulate Jacobi iterative method to solve the set of linear equations. A group of learnable class logits $a_i(i=1,2,...,n+1)$ are used to estimate the weights $w_i$ with a softmax function,
\begin{equation}
    w_i = \frac{(n+1) \times e^{a_i}}{\sum_{j=1}^{n+1}e^{a_j}}.
    \label{eql:logits}
\end{equation}
The softmax function here can ensure the value of $w_i$ to be positive and the $a_i$ are initially set to 0 at the beginning of training. Denote the weights at the $t$th iteration as $w^t_i$. Then we update the $w_i$ toward a target weight $\hat{w}^{t+1}_i$ to obtain $w^{t+1}_i$. The $\hat{w}^{t+1}_i$ are calculated by Jacobi iterative method,
\begin{equation}
    \hat{w}^{t+1}_i = (-\sum_{k \neq i} w^t_k \times   \boldsymbol{\tilde G}_{ik}) /\boldsymbol{\tilde G}_{ii}.
    \label{eql:jacobi}
\end{equation}
For background class, $ \hat{w}^{t+1}_{n+1} = n + 1 - \sum_{i=1}^{n}\hat{w}^{t+1}_{i} $. 
Then, a new loss is designed to aligned the $w^{t+1}_i$ with $\hat{w}^{t+1}_i$ by normalizing $a_i$ at $t$th iteration,
\begin{equation}
    L_{align} = \frac{1}{n+1} \sum_{i=1}^{n+1} (log(\hat{w}^{t+1}_i) - a_i)^{2},
\end{equation}
where the mean square loss is applied on the logits aspect to prevent the loss dominated by the large weights.

Finally, the classification losses for labeled and unlabeled images are summarized as follows,
\begin{equation}
    L = \sum_{i=1}^{n+1} w_i^l \times \sum_{c_j \in C_i}{f^l(x_j)} + \sum_{i=1}^{n+1} w_i \times \sum_{c_j \in C_i}{f^u(x_j)} + L_{align}
\end{equation}
Gradient-based reweighting alleviates the imbalance from the gradient-level aspect. It is complementary to the class-rebalancing sampling as it is beneficial to classes with easy samples and classes with a low occurrence frequency in the unlabeled set.

\subsection{Gradient-based Thresholding}
\label{sec:GbT}
\begin{figure}[ht]
    \centering
    \includegraphics[clip,trim={0 0 1cm 0}, width=0.8\linewidth]{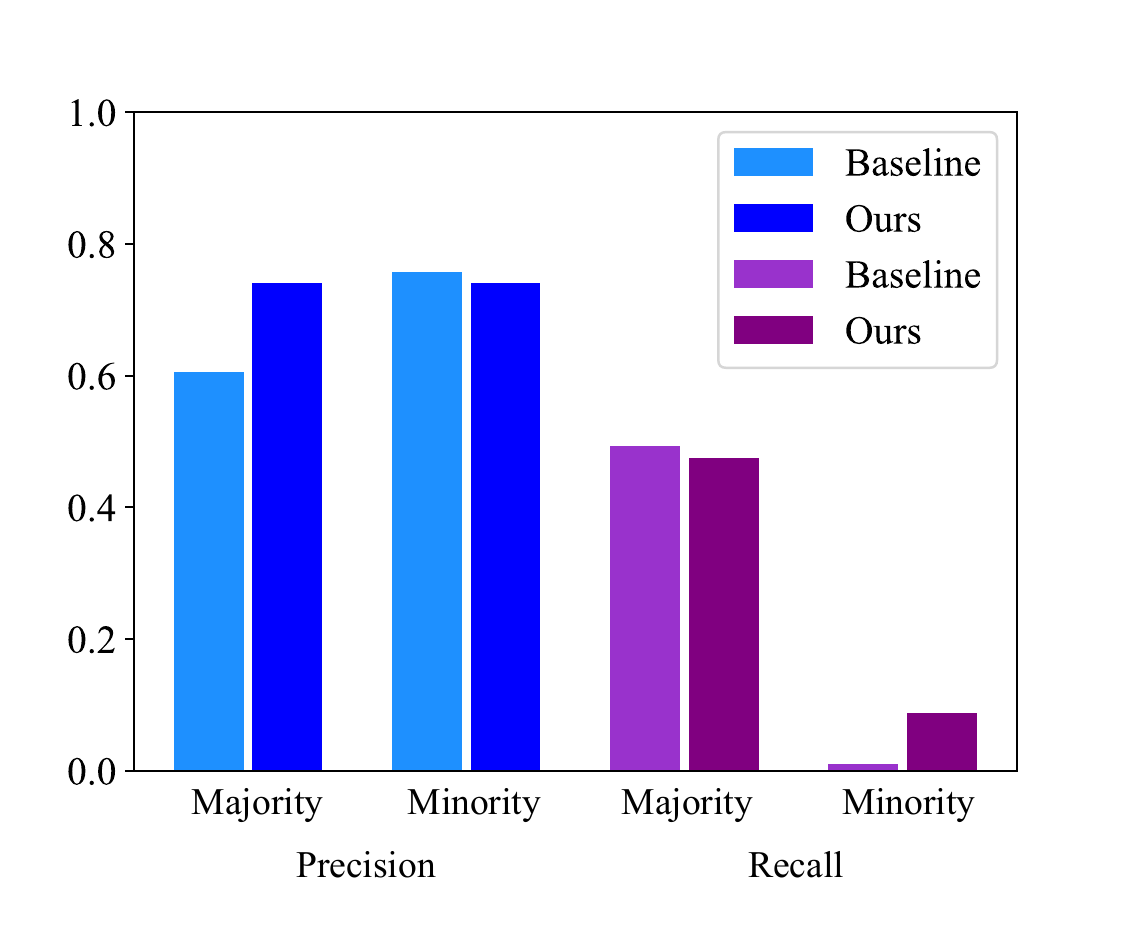}
    
    \caption{Comparison of precision and recall of pseudo labels in Soft teacher baseline and our proposed method. Our method generates more pseudo labels for minority classes while preserving the precision of all classes, thus achieving more balanced total detection performance.
    }
    \label{fig:precision}
    \end{figure}

Current SSOD methods typically employ a fixed threshold $\theta$ to sample pseudo labels. However, we observe that the precisions of minority classes under a fixed threshold are relatively high and the recall of them is extremely low as presented in Fig.~\ref{fig:precision}. Clearly, this fixed threshold strategy makes it hard for the object detector to generate enough pseudo labels for minority classes. Besides, dynamic thresholding mechanisms\cite{kaul2022label} are also proposed to cope with this issue, however, they are mainly based on the distributions of scores and fail to consider the complexity of sample space during model training. To mitigate this issue, we estimate the model bias from the gradients imbalance perspective. Specifically,
we devise a gradient-based adaptive thresholding strategy to decide the class-wise thresholds. 
Concretely, we utilize the aforementioned class-wise balancing weights to generate thresholds for class $C_i$, 
\begin{equation}
\theta^p_i = min(\theta, \frac{\theta}{w_i}).
\end{equation}
Obviously, $\theta_i^p$ has a smaller value for classes with higher ${w_i}$. This enables the object detector to adaptively harvest more pseudo labels for those minority classes, thereby alleviating the class imbalance problem via solving the first type of confirmation bias. Complementary to the GbR module, it prevents detectors from overfitting with the small number of pseudo labels for minority classes, especially when the class weights of minority classes are larger. We also incorporate a score-based thresholding strategy\cite{nie2023adapting} to obtain the final thresholds $\theta_i$ by $\theta_i = min(\theta_i^p, \theta_i^c)$ to achieve better performance, where $\theta_i^c$ is the thresholds estimated from \cite{nie2023adapting}.
.

\subsection{Class-rebalancing Sampling}
\label{sec:CrS}
The aim of this module is to alleviate the influence of incorrect pseudo labels which causes the second type of confirmation bias.
Particularly, when applying the LTOD methods on unlabeled data, a significant number of incorrect pseudo labels for minority classes are generated. The incorrect pseudo labels with resampling or reweighting in these methods dominate the learning of the detector for minority classes, especially when the number of labeled data is few. We further analyze this issue in the Supplementary Document.
To cope with this issue, we perform a score-based resampling. Concretely, we first conduct data resampling with Repeat factor sampling(RFS)~\cite{gupta2019lvis} from the training set $\boldsymbol{D}_l$. Denote the unlabeled data $\boldsymbol{D}_u$ with its pseudo labels at generation $t$ as $\boldsymbol{D}_u^t$. We estimate the resampling rate based on the probability score and frequency of $\boldsymbol{D}_u^t $ with class-wise thresholds from the GbT module.

To illustrate the adaptive class-wise sampling in each generation $t$, let the number of images containing at least one box annotation for class $C_i$ in $\boldsymbol{D}_u^t \cup \boldsymbol{D}_l$ be $m_i$.  We expect that the sampling rates are large for minority classes and small for majority classes. Similar to RFS\cite{gupta2019lvis}, for each class $C_i$, the repeat sampling rate is defined as,
\begin{equation}
S_{i} = max(1, \sqrt{\frac{\epsilon \times (N_l + N_u)}{m_i}}).
\end{equation}
$\epsilon = \gamma \times t/N_t$ is an adaptive sampling factor rising during the training process to learn better representations\cite{he2021rethinking}.
Here $\gamma$ is the coefficient to modify the range of sampling rates and $N_t$ is the number of total generations.

Next, suppose the pseudo labels of an unlabeled image $\boldsymbol{I}^u_i$ is set to ${(c^u_{ij}, p^u_{ij}, b^u_{ij})}_{j=1}^{N_i}$, where $c^u_{ij}$, $p^u_{ij}$ and $b^u_{ij}$  denote the category, probability scores and bounding boxes positions of pseudo labels, respectively. $N_i$ is number of pseudo labels in $\boldsymbol{I}^u_i$. 
Different from sampling rates in \cite{gupta2019lvis}, for each pseudo ground-truth label $(c^u_{ij}, p^u_{ij}, b^u_{ij})$, we estimate a sampling rate based on scores of pseudo labels to calculate the repeat times of this image,
\begin{equation}
S^u_{ij} = S_{c^u_{ij}} (p^u_{ij} - \theta_{c^u_{ij}}).
\end{equation}
 $\theta_{c^u_{ij}}$ here is the adaptive threshold of class $c^u_{ij}$. The sampling rate of image $\boldsymbol{I}^u_i$ are defined as the max value in ${S^u_{ij}}_{j=1}^{N_i}$. With this sampling rate, the model tends to learn the pseudo labels with high confidence, alleviating the disturbance from incorrect pseudo labels, especially for minority classes, which helps to tackle the second type of confirmation bias. Since the sampling rate is also determined by $\theta_{i}$ in the GbT module, the CrS module smooths the gradients for both the GbR and GbT modules and keeps the learning stable by sampling more times for classes with small weights.

\section{Experiments}
\subsection{Datasets}
\begin{figure}[ht]
    \centering
    \includegraphics[clip,trim={0 0 0 0}, width=\linewidth]{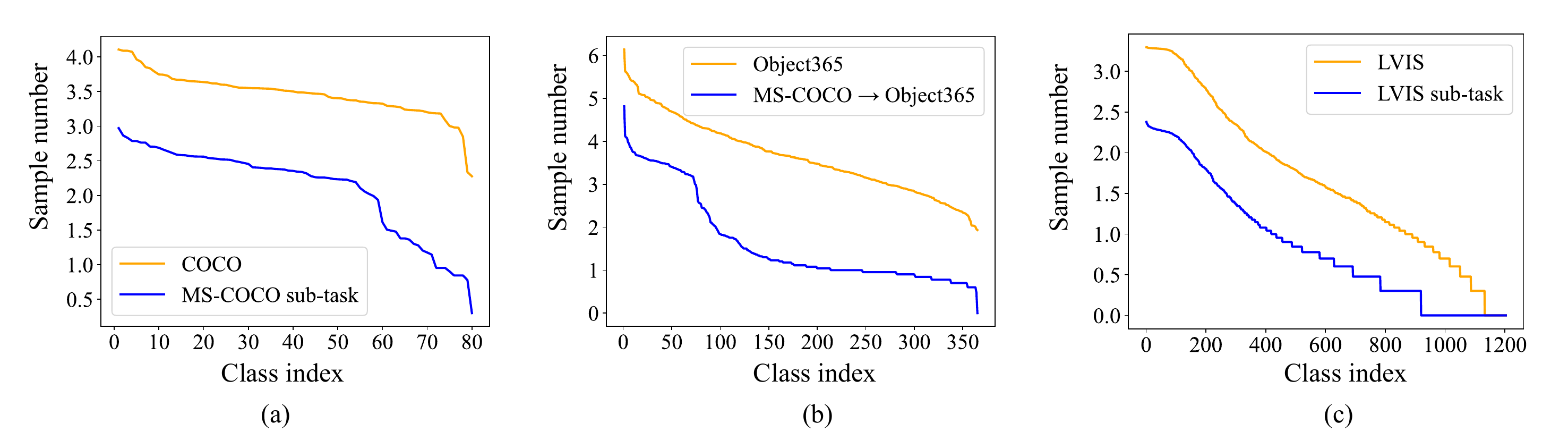}
    
    \caption{Sample distributions scaled by log10 of all classes in original dataset v.s. our proposed sub-task. (a) COCO v.s. COCO sub-task; (b) Object365 v.s. MS-COCO $\rightarrow$ Object365 sub-task; (c) LVIS v.s. LVIS sub-task. Note that the `class index' is sorted by class frequency and does not correspond to the actual class number.}
    
    \label{fig:difference_coco}
    \end{figure}

 As introduced in Section~\ref{sec:def}, we introduce two scenarios with different distributions in unlabeled data for CI-SSOD. In the first scenario, we introduce the MS-COCO task and the MS-COCO $\rightarrow$ Object365 task. In the second scenario, we introduce the LVIS sub task. Here, we visualize the sample distributions (scaled by log10) of all classes in the three sub-tasks shown in Fig.~\ref{fig:difference_coco}. The distribution of the unlabeled data generally follows the distribution of the benchmark dataset. Specifically, in MS-COCO and MS-COCO $\rightarrow$ Object365 sub-tasks, there are abundant unlabeled instances for minority classes in the unlabeled data, and the class imbalance of the labeled data is more severe than that of the unlabeled data. In the LVIS sub-task, the instances of minority classes in unlabeled data are naturally scarce, and the distribution of labeled and unlabeled data is seriously class imbalanced.

\textbf{MS-COCO~\cite{lin2014microsoft}} is a dataset for object detection, which consists of 80 classes. For the proposed MS-COCO sub-task, we evaluate our methods on two splits. In the first split, we select 20 classes shared with Pascal VOC\cite{everingham2010pascal} as the minority classes and other classes are set as the majority classes. In the second split, we randomly split for 40 classes as minority classes while the other 40 classes are split as majority classes. In MS-COCO sub-task, the labeled set $\boldsymbol{D}_l$ consists of a large number of samples for the majority classes and a small number of samples for the minority classes. Specifically, we construct the $\boldsymbol{D}_m$ for the majority classes by sampling 10\% images from all the images containing the majority classes $\boldsymbol{C}^f$. For the minority dataset $\boldsymbol{D}_s$, we ensure at least 10 ground-truth bounding box annotations are collected for each minority class in $\boldsymbol{C}^r$. 
Different from the setting in FSOD, to prevent information leakage, we annotate all instances within an image in our tasks. Specifically, when constructing $\boldsymbol{D}_m$, we randomly select images for minority classes until each class has a minimum of 10 instances. Some selected images may exceed this threshold due to coexistence with other minority classes. 
The remaining images that are not sampled in $\boldsymbol{D}_l$ in MS-COCO are set as unlabeled set $\boldsymbol{D}_u$. The performance of detectors is evaluated on 5K images from COCO2017 validation set similar to the standard MS-COCO detection metrics. 

\textbf{Object365~\cite{shao2019objects365}} is a more challenging dataset with 175k images and 365 classes. In our proposed MS-COCO $\rightarrow$ Object365 sub-task, we combine these two standard benchmark datasets together to better mimic real-world scenarios. Specifically, images in MS-COCO are fully annotated with 78 identical classes shared with Object365, which is set to be the majority class dataset $\boldsymbol{D}_m$. In order to match the classes in Object365, ``Teddy bear'' in MS-COCO is set as ``bear'' and ``sports ball'' is eliminated. The remaining 287 classes are regarded as the minority classes. Similar to the MS-COCO sub-task, at least 10 ground-truth bounding box annotations are labeled to each minority class. The remaining images of Object365 are regarded as $\boldsymbol{D}_u$. The performance of a detector is evaluated based on the validation set of Object365 with 80k images.

\textbf{LVIS~\cite{gupta2019lvis}}(v1.0) is a long-tailed dataset with reannotated images from MS-COCO for 1203 classes. LVIS consists of a training set with almost 100k images and a validation set with 20k images. In our proposed LVIS sub-task, 10\% of the training set is randomly selected as $D_l$ under the premise that at least 1 ground-truth bounding box annotations for each class are contained in $D_l$. The other images in the training set are set as $D_u$. We evaluate the detectors on the validation set of LVIS to demonstrate that our method performs well when there are few or no implicit instances for some minority classes in $D_u$.

\subsection{Implementation Details}
 Following \cite{xu2021end}, we employ the Faster-RCNN~\cite{ren2015faster} with ResNet-50~\cite{He_2016_CVPR}  and FPN~\cite{Lin_2017_CVPR} as our baseline object detector. The object detector is trained on 8 GPUs with batch size of 5 per GPU, 1 for labeled images and 4 for unlabeled images. The learning rate is set to 0.01 initially and is divided by 10 at 75\% and 90\% of total iterations. 
 We train detectors for 10 generations for both datasets, with 180k iterations for the MS-COCO task and LVIS task and 360k iterations for the MS-COCO $\rightarrow$ Object365 task.
 SGD~\cite{ruder2016overview} is selected as our optimizer with the momentum set to 0.9 and the weight decay set to 1e-4. Focal loss\cite{lin2017focal} is applied in classification loss for supervised and unsupervised images. $\gamma$ is set to 0.5 for MS-COCO task and 0.05 for other tasks. $\beta$ is set to 0.5.
Following \cite{xu2021end}, $\theta$ is set to 0.9. The moving average coefficients $\eta_p$ and $\eta_g$ are both set to 0.9995. 

\begin{table}[]
    \centering
    \caption{Quantitative comparison with state-of-the-art methods on the MS-COCO sub-task. All results are shown in average precision(AP) in standard coco metric. ``all'', ``maj'' and ``min'' denote the AP of all classes, majority classes and minority classes, respectively.}
    \resizebox{\linewidth}{!}{
    \begin{tabular}{l|lll|lll}
    \hline 
    \multicolumn{1}{c|}{\multirow{2}{*}{Method}} & \multicolumn{3}{c|}{split1} & \multicolumn{3}{c}{split2} \\ \cline{2-7} 
    \multicolumn{1}{c|}{}                        & all     & maj     & min     & all     & maj     & min     \\ \hline \hline
    Soft teacher\cite{xu2021end}  & 23.6  &  31.0  &  1.5   & 17.2   &  32.7  &  1.7  \\ 
    RFS\cite{gupta2019lvis}  &  23.9  &  30.2   &  4.8   &  19.8   &  32.3   &  7.2  \\ 
    Eqlv2 \cite{tan2021equalization} & 22.8  &  29.1 &  6.5 &  20.2   &  31.4  &  9.1  \\
    C2AM Loss\cite{wang2022c2am} &  24.4   &  30.8  &   6.2  & 20.8  &  32.9 &  8.6  \\ 
    CReST\cite{wei2021crest} &   22.2    &  29.3   &  1.0    & 16.3  & 31.2 &  1.4  \\ 
    DASO\cite{oh2022daso}  & 24.8  &  \textbf{31.1}  & 6.0 & 20.7 & \textbf{33.0}  &  8.4 \\
    DeFRCN\cite{qiao2021defrcn}  &  24.4 &   29.4 &  9.2  &  20.9  &  30.5  &  11.3  \\ 
    \hline
    Ours & \textbf{26.5}   &   30.7    &  \textbf{13.9}  & \textbf{23.8} & 32.4 & \textbf{15.1} \\
    \hline
    \end{tabular}}
    \label{tab:coco}
\end{table}

\begin{table}[]
    \centering
     \caption{Quantitative comparison with state-of-the-art methods on the MS-COCO $\rightarrow$ Object365 sub-task. All results are shown in average precision(AP) in standard coco metric.}
    \begin{tabular}{l|ccc}
    \hline
    Method &  all & maj & min \\ \hline
     Soft teacher\cite{xu2021end} &   4.8   &   22.1    & 0.2         \\
    RFS\cite{gupta2019lvis} & 7.2 &  23.1  &   2.9       \\ 
     Eqlv2 \cite{tan2021equalization} &   7.4   &    20.3   &   3.9    \\
     C2AM \cite{wang2022c2am}  &  7.6    &   23.2  &   3.4    \\
     CReST\cite{wei2021crest} &   6.8    &   21.3    &   2.8   \\ 
     DASO\cite{oh2022daso}   &   7.8   &   23.3   &  3.6    \\ 
    DeFRCN\cite{qiao2021defrcn} &  7.9    &  21.2   &  4.3    \\
     \hline
     Ours &  \textbf{9.2}   &   \textbf{23.7}    &  \textbf{5.2}  \\
     \hline
    \end{tabular}
    \label{tab:object365}
    \end{table}

\subsection{Comparison to State-of-the-art Methods}
We compare our method against 7 state-of-the-art methods on two experiment settings in Tab.~\ref{tab:coco} and Tab.~\ref{tab:object365}, including Soft teacher\cite{xu2021end} for SSOD, RFS\cite{gupta2019lvis}, Eqlv2\cite{tan2021equalization} and C2AM\cite{wang2022c2am} for LTOD, CReST\cite{wei2021crest}, DASO\cite{oh2022daso} for class imbalanced semi-supervised image classification and DeFRCN\cite{qiao2021defrcn} for FSOD. Note that Soft-teacher is selected as our baseline method. Generally, we combine detectors in FSOD, LTOD, and semi-supervised image classification with SSOD (Soft-teacher). The sampling strategies in these methods are replaced with RFS. For LTOD, we replace the classification loss with losses in these methods(Eqlv2, C2AM). For DeFRCN, we pretrain a detector for the majority classes and fine-tune the detector in a balanced dataset with both majority and minority classes.

For the MS-COCO sub-task shown in Tab.~\ref{tab:coco}, our method surpasses all these methods significantly under Average Precision(AP) in both splits. Concretely, in split 1, our method outperforms the second-best method DASO by 1.5\% and beats the baseline Soft teacher by 2.7\%. Interestingly, it is clear that current SSOD methods, for example, Soft teacher in row 1, generate inferior performance in our proposed experiment setting, achieving 31.0\% on the majority classes and 1.5\% on the minority classes. This strengthens our observation that current SSOD detectors are not able to perform well under challenging class-imbalanced datasets.
However, our method boosts the performance of minority classes from 1.5\% to 13.9\%. For minority classes, our method beats the second-best performing method Defrcn by 4.7\%. In split 2, the performance of our method surpasses that of the baseline by 6.6\%. Our method also defeats the DeFRCN method by 3.8\% in minority classes and 2.9\% in total. The comparisons show that the combination of the methods with SSOD could alleviate the class imbalance to some extent compared to the Soft teacher baseline. However, they suffer from overfitting and the two types of confirmation biases. In contrast, our method can effectively alleviate confirmation biases. 

For the MS-COCO $\rightarrow$ Object365 sub-task shown in Tab.~\ref{tab:object365}, the AP of our method is higher than DeFRCN by 1.3\% on all classes and by 0.9\% on minority classes. Our method obtains an improvement of 0.6\% since there exists a small imbalance among the majority classes. The more challenging experiment setting further validates the effectiveness of our method to tackle class imbalance and confirmation biases in extremely class-imbalanced datasets.
\begin{table}[]
    \centering
     \caption{Quantitative comparison with state-of-the-art methods on the LVIS sub-task. All results are shown in average precision(AP) in standard coco metric.  ``f'', ``c'' and ``r'' denote the AP of frequent classes, common classes and rare classes.}
    \begin{tabular}{l|cccc}
    \hline
    Method &  all & f & c & r \\ \hline
     Soft teacher\cite{xu2021end} &   7.6   &   22.5    & 11.9 & 2.0\\
    RFS\cite{gupta2019lvis} & 8.0 &  23.1  &   12.7   &  2.3  \\ 
     Eqlv2 \cite{tan2021equalization} &   8.1   &    20.2   &   11.8  &  3.5 \\
     C2AM \cite{wang2022c2am}  &  8.6    &   22.1   &   12.3  &  3.7 \\
     DASO\cite{oh2022daso}   &   8.7   & \textbf{23.2}  & \textbf{13.1}   &  3.2   \\ 
     \hline
     Ours &  \textbf{8.9}   &   21.7    &  12.3  & \textbf{4.2} \\
     \hline
    \end{tabular}
    \label{tab:lvis}
    \end{table}

For the LVIS sub-task shown in Tab.~\ref{tab:lvis}, we report the results against other methods on classes with different frequencies in our selected training set. Following previous LTOD methods, the classes are split into rare classes($\leq$ 10 images), common classes(11-100 images), and frequent classes($>$ 100 images). The detectors suffer more from the second type of confirmation bias than detectors in other tasks since there are few (usually $<$ 10) or even no implicit instances for minority classes in unlabeled images. Our method outperforms the best LTOD method by 0.2\% in all classes and 0.5\% in rare classes, demonstrating superiority in tackling the second type of confirmation bias.

\subsection{Ablation Study}
We tease apart critical algorithmic components or apply different settings for them, forming variants of our method.  Tab.~\ref{tab:ablation} shows the AP of variants of our method. It can be observed that the detector gains substantial improvement of 5.3\%, 4.2\%, 6.4\% with CrS, GbT, GbR on minority classes while the performance of majority classes levels off. It shows our modules can effectively enhance the capability of detectors for minority classes without a substantial drop for majority classes. With the focal loss, the performance of the detector obtains a considerable increase of 2\% but only slightly enhances by 0.2\% on all classes. When we eliminate CrS, GbT, GbR in our method, the performance drops by 3.4\% , 2.5\%, 3.6\% for minority classes and 0.7\%, 0.7\%, 0.8\%for all classes. 

To further validate the effectiveness of our gradient-based thresholding mechanism, we conduct experiments under different thresholds and present the results in Tab.~\ref{tab:thresholds}. Our gradient-based thresholding module outperforms the fixed threshold at 0.5 by 0.9\% and at 0.9 by 0.7\%.  The performance of our method is higher than that with only one threshold module $\theta^c_i$ or $\theta^p_i$, demonstrating the complementary of our proposed thresholds. The performance of our method also surpasses existing thresholding in LabelMatch~\cite{Chen_2022_CVPR} in SSOD. This is because LabelMatch is based on the premise that labeled and unlabeled data share the same class distribution, which is not satisfied in the MS-COCO sub-task.  In contrast, our thresholding mechanism can generate more high-quality pseudo labels to tackle the confirmation bias with a high generalization capability under different scenes.

\begin{table}[]
    \centering
    \caption{Comparison of our core components on the MS-COCO sub-task under Average Precision(AP). ``CrS'' denotes Class-rebalancing sampling. ``GbT'' denotes Gradient-based thresholding) and ``GbR'' denotes Gradient-based reweighting.  ``FL'' denotes the losses for classification are replaced with Focal loss. }
    \begin{tabular}{ccc|ccc}
    \hline
     CrS  &  GbT & GbR   & all & maj & min \\ \hline \hline
    &  & & 23.6 &    31.0 & 1.5         \\ 
    &  & FL & 23.8 &    30.5 & 3.5         \\ 
    \checkmark & & & 24.6  & 30.7 & 6.6 \\
    & \checkmark & & 24.8 & \textbf{31.2} & 5.7   \\ 
     & & \checkmark & 25.2 & 31.0 & 7.9  \\ 
     & \checkmark &\checkmark & 25.8&  30.9  &  10.5 \\ 
     \checkmark &  & \checkmark & 25.8  &  30.6  &  11.4   \\ 
    \checkmark & \checkmark & & 25.6  &   30.8  & 10.3   \\ 
    \checkmark & \checkmark & \checkmark & \textbf{26.5}  & 30.7 & \textbf{13.9} \\ \hline
    \end{tabular}
    
    \label{tab:ablation}
    \end{table}

    \begin{table}[]
    \centering
    \caption{Comparison of performance under different thresholds on the MS-COCO sub-task under AP.  }
    \begin{tabular}{c|cccc}
    \hline
      & thres & all & maj & min \\ \hline
    1 & 0.5 & 25.6  &  30.0   &     12.6   \\ \hline
    2 & 0.9 & 25.8  & 30.6   &  11.4   \\ \hline
    3 & $\theta^c_i$ & 25.9 & \textbf{30.8}  & 11.9 \\ \hline
    4 & $\theta^p_i$ & 26.4 & 30.6 & 13.8 \\ \hline
    3 & LabelMatch & 25.9  & \textbf{30.8}   &  12.1 \\ \hline
    4 & GbT & \textbf{26.5} &  30.7 &  \textbf{13.9}         \\ \hline
    
    \end{tabular}
    
    \label{tab:thresholds}
    \end{table}

\section{Conclusion}
We analyzed the class imbalance problem in current SSOD task and identified a prominent research gap of the current SSOD experiment setting. To close this gap, we presented the first experiment setting for class imbalanced semi-supervised object detection. Moreover, we introduced a simple yet effective gradient-based sampling framework to deal with the class imbalance from the perspective of two confirmation biases. We extensively verify the effectiveness of our proposed components and the results demonstrate that our method outperforms current class imbalance based methods by clear margins. Our task setting and method could serve as baselines for future CI-SSOD research.

\section*{Acknowledgments}
This work was supported in part by the Guangdong Basic and Applied Basic Research Foundation (NO.~2020B1515020048), in part by the National Natural Science Foundation of China (NO.~61976250), in part by the Shenzhen Science and Technology Program (NO.~JCYJ20220530141211024).

\section*{Appendix}
Please refer to our conference version.

{\small
\bibliographystyle{ieee_fullname}
\bibliography{egbib}
}

\end{document}